\documentclass{article}

\usepackage{arxiv_default}
\usepackage[utf8]{inputenc}
\usepackage[T1]{fontenc}
\usepackage{hyperref}
\usepackage{url}
\usepackage{booktabs}
\usepackage{amsfonts}
\usepackage{amsmath}
\usepackage{amssymb}
\usepackage{nicefrac}
\usepackage{microtype}
\usepackage{xcolor}
\usepackage{algorithm}
\usepackage{algorithmic}
\usepackage{multirow}
\usepackage{graphicx}
\graphicspath{{figures/}}
\usepackage{subcaption}
\usepackage{placeins}
\usepackage{amsthm}

\usepackage{enumitem}

\newcommand{\ours}{\textsc{CoCommit}}
\newcommand{\mask}{\texttt{[M]}}
\newcommand{\MtoT}{\textrm{M2T}}
\newcommand{\TtoT}{\textrm{T2T}}
\newcommand{\ctx}{\mathrm{ctx}}
\newcommand{\TC}{\mathrm{TC}}
\newcommand{\dset}{\mathcal{D}}           % decision set
\newcommand{\cset}{S}                     % commit bundle
\newcommand{\marker}{\mathbf{m}}          % marker vector
\newcommand{\Mset}{\mathcal{M}}           % masked positions (training)
\newcommand{\Wset}{\mathcal{W}}           % wrong-visible positions (training)
\newcommand{\Gset}{\mathcal{G}}           % clean-visible positions (training)

\title{Don't Commit Alone: Joint Token Commitment in Diffusion Language Models}

\author{
  Lin Yao$^{1,2}$ \\
  $^1$School of Computer Science, Shanghai Jiao Tong University, Shanghai, 200240, China \\
  $^2$Zhongguancun Academy, Beijing, 100097, China \\
  \texttt{lin.yao@sjtu.edu.cn}
}

\begin{document}

\maketitle
% !TEX root = main.tex
% =====================================================================
% body.tex -- full paper body.
% =====================================================================

\begin{abstract}
Diffusion language models (dLLMs) commit multiple tokens per denoising step by decoding each selected position independently from a shared context. When these positions are dependent, this factorization introduces an error captured by conditional total correlation, which confidence-based selection cannot infer from marginal probabilities alone. We propose \ours{}, a marker-gated coordination pass that delays commitment. After the usual bundle selection, a learned marker identifies the commit set, and the backbone's last $n$ layers are re-applied to coordinate the marked positions before greedy argmax writes the tokens. This approximates joint-mode decoding while reusing existing weights, requiring only one partial forward pass and no auxiliary model. On LLaDA2.1-mini with LoRA adapters and greedy inference, joint commitment improves five of the seven evaluated benchmarks over the released factorized decoder. The largest gains occur on code and reasoning tasks, while the remaining tasks are near parity.
\end{abstract}

%=============================================================================
\section{Introduction}
\label{sec:intro}
%=============================================================================

Diffusion language models (dLLMs) generate text by iteratively revealing tokens from a masked sequence, committing multiple positions in parallel at each denoising step~\citep{austin2021d3pm,sahoo2024mdlm,lou2024sedd,nie2025llada,ye2025dream}. Parallelism is the source of their speed advantage over autoregressive decoding, and also the source of a characteristic error: tokens committed in the same step are predicted from the same context but not from each other, so each is drawn from its own marginal distribution while the tuple they form may belong to no coherent joint completion~\citep{kang2025parallelbench}.

This failure is often described as an operational problem: parallel tokens can be mutually inconsistent. Existing methods address it downstream through post-hoc repair. Token-to-token editing~\citep{nie2026llada21} and remasking~\citep{wang2025remdm,yao2026remask} overwrite or redraw committed tokens in later steps. Each redraw, however, remains a factorized prediction. Repair repeatedly applies the same marginal readout that produced the inconsistency, either sequentially or in an ICM-style procedure~\citep{besag1986icm}, and each iteration incurs a full forward pass. It never predicts the dependent positions jointly, leaving the original commitment rule unchanged. We instead characterize the error directly. For a commit set $\cset$ and context $\ctx$, factorized commitment replaces the joint posterior $p(x_\cset \mid \ctx)$ with the product of marginals $\prod_{i \in \cset} p(x_i \mid \ctx)$. Their discrepancy is the conditional total correlation $\TC(x_\cset \mid \ctx)$, which vanishes if and only if the committed positions are conditionally independent given the context. This quantity governs the quality--speed trade-off in parallel decoding. Autoregressive decoding obtains exactness through sequentiality ($|\cset| = 1$, $\TC \equiv 0$), whereas parallel commitment pays $\TC(x_\cset \mid \ctx)$ at each step in exchange for speed.

This formulation reveals a limitation of current commit-set selection. Confidence-based rules select positions with the most concentrated marginals, so they measure marginal entropy rather than dependence. Marginals average over the joint modes of the posterior. For example, in a context compatible with both ``New York City'' and ``San Francisco'', the first position may assign high probability to both \emph{New} and \emph{San}, while different positions derive their confidence from different modes. High marginal confidence therefore does not imply low total correlation, and a rule that observes only marginals cannot bound the commitment error. Better threshold tuning cannot recover dependency information that is absent from the observed quantities.

If selection cannot avoid dependent bundles, the commitment operation must handle them. The backbone's hidden states already encode the joint structure of the block, but the output interface discards this information. Its factorized readout head projects each position onto the vocabulary independently when the positions are committed. The missing component is therefore a \emph{communication channel at decision time}, rather than additional model capacity.

\begin{figure}[t]
    \centering
    \includegraphics[width=\linewidth]{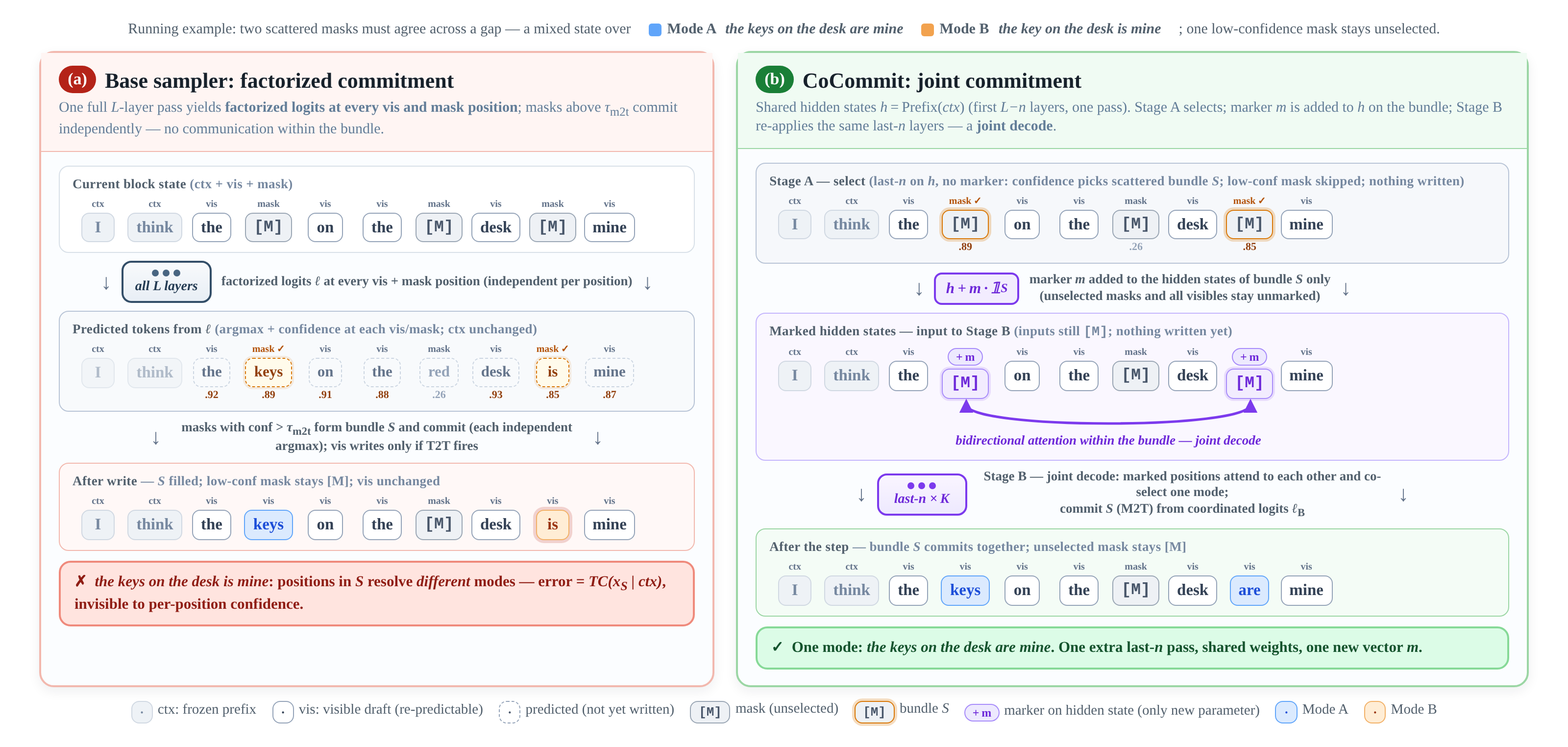}
    \caption{Factorized versus joint commitment (one denoising step).
    \textbf{(a)~Base sampler:} a single $L$-layer pass yields factorized logits at every visible and masked position. Masks above $\tau_{\mathrm{m2t}}$ form bundle $\cset$ and commit by independent argmax. Positions in $\cset$ can resolve different joint modes (e.g.\ \emph{the keys \ldots\ the desk is mine}) even when every marginal is confident; unselected masks stay \mask{}.
    \textbf{(b)~\ours{}:} shared prefix states $h=\mathrm{Prefix}(\ctx, z)$; Stage~A selects $\cset$ from unmarked factorized logits; marker $\marker$ is added to $h$ on $\cset$ only; Stage~B re-applies the same last-$n$ layers for a joint decode before M2T/T2T writes.}
    \label{fig:overview}
\end{figure}

\ours{} adds this channel with a small modification (Figure~\ref{fig:overview}). Each denoising step is split into two stages that share a prefix computation. Stage~A runs the last $n$ transformer layers without modification and selects the commit bundle using the standard confidence rule, but does not write tokens. A learned marker vector is then added to the hidden states of the selected positions. Stage~B re-applies the same last $n$ layers, allowing attention among the marked positions to reduce mode averaging before commitment. The final tokens are read from the Stage~B logits. Under greedy commitment, deterministic coordination followed by per-position argmax tends to produce a tuple from one joint mode rather than a mixture of modes. The target is thus the joint \emph{mode} of $p(x_\cset \mid \ctx)$, not samples from the joint distribution. The readout remains one softmax per position, and this interpretation is limited to greedy argmax. With temperature sampling, independent noise may still split modes; the model may also commit an incorrect but self-consistent mode. Post-hoc repair methods address these complementary errors. During training, branch~2 predicts the marked positions conditioned on the marker. Edit and clean losses constrain predictions at unmarked positions, so the marker can help only through attention among marked positions.

The mechanism extends along two axes. First, the \emph{decision set} need not be restricted to masked positions. Marking a visible token would submit its current value for joint reconsideration, unifying commitment (mask-to-token) and revision (token-to-token) within one operation. Our experiments mark only masked commit bundles. Second, the coordination pass can be iterated. Setting $K{=}0$ recovers factorized decoding, while $K{=}1$ is our main setting. With injected noise and denoising training, deeper iterations could trade test-time compute for stronger within-bundle coordination and approach a conditional joint model in the spirit of MAR~\citep{li2024mar}.

dLLMs exhibit two distinct errors. \emph{Within-step} factorization assembles potentially incoherent tuples from confident marginals, with error $\TC(x_\cset \mid \ctx)$. \emph{Across-step} semantic error occurs when an early incorrect commit steers later predictions as context. \ours{} addresses the first error at commitment time, while repair methods address the second. The two approaches are complementary.

This paper makes the following contributions:
\begin{itemize}[nosep,leftmargin=*]
    \item We motivate joint token commitment by framing within-step parallel error as conditional total correlation of the commit bundle, and by identifying a blind spot in confidence-based selection: per-position marginals average over joint modes and carry no dependency signal.
    \item We propose \ours{}, a marker-gated coordination pass that reuses the backbone's last $n$ layers so the commit bundle exchanges information before token writes. It requires no new prediction head or external dependency model and adds one partial forward pass (Section~\ref{sec:method}).
    \item We evaluate \ours{} on seven standard benchmarks spanning math, science, reading comprehension, commonsense, and code. We train only LoRA adapters and marker vectors on generic prose. Joint commitment improves five tasks, with the largest gains on code and reasoning, and remains within $0.3$ points on the other two (Table~\ref{tab:main}).
\end{itemize}

%=============================================================================
\section{Related Work}
\label{sec:related}
%=============================================================================

\paragraph{Discrete diffusion language models.}
D3PM~\citep{austin2021d3pm} established diffusion over discrete tokens and identified absorbing-state (masking) corruption as best suited to text; SEDD~\citep{lou2024sedd}, MDLM~\citep{sahoo2024mdlm}, and MD4~\citep{shi2024md4} progressively simplified the formulation, and RADD~\citep{ou2025radd} showed that the learned predictor is exactly a conditional distribution over clean tokens given the unmasked context. LLaDA~\citep{nie2025llada} and its successors~\citep{nie2026illada,nie2026llada21} scaled the recipe to large models, Dream~\citep{ye2025dream} and DiffuLLaMA~\citep{gong2025diffullama} adapted autoregressive checkpoints, and BD3-LM~\citep{arriola2025bd3lm} introduced the semi-autoregressive block schedule that LLaDA2.1~\citep{nie2026llada21} adopts. These methods commit multiple positions per step by independent argmax or sampling from per-position distributions. Our work changes this commitment rule while leaving the rest of the pipeline unchanged.

\paragraph{Continuous diffusion language models.}
A parallel line keeps the generative trajectory in a continuous space and discretizes late. ELF~\citep{hu2026elf} denoises in a frozen encoder's embedding space with flow matching and maps to tokens only at the terminal step; Cola~\citep{guo2026cola} transports a latent prior with a block-causal DiT and decodes text from the latent. In our vocabulary, these models coordinate positions continuously throughout the trajectory and never perform a mid-trajectory parallel factorized commit; the price is the loss of early discrete pruning and a terminal continuous-to-discrete mapping step. We treat this family as the late-discretization reference point of the design space, not as a competitor: \ours{} localizes ELF-style pre-discretization coordination to each commit instant of an early-committing model, rather than returning to fully late discretization.

\paragraph{Dependency-aware parallel decoding.}
The inconsistency of independently decoded parallel tokens predates dLLMs: non-autoregressive translation observed it as the multimodality problem~\citep{gu2018nat}, and iterative refinement schemes such as Mask-Predict~\citep{ghazvininejad2019maskpredict} amortize it across steps. ParallelBench~\citep{kang2025parallelbench} documents the same failure in modern dLLMs. Closest to our diagnosis, discrete copula diffusion~\citep{liu2024copula} shows that per-position marginals miss the dependency structure (the copula) of the joint and restores it with a separate autoregressive model at inference. We internalize the correction instead: the backbone itself is made dependency-aware at commit time through a marker-gated re-application of its own last-$n$ layers, requiring no auxiliary model and matching the greedy-commitment regime these systems actually use.

\paragraph{Repair-based self-correction.}
An orthogonal family repairs commitments after they are written: token-editing phases in LLaDA2.1~\citep{nie2026llada21}, uniform or targeted remasking~\citep{wang2025remdm,zhai2026core,yao2026remask}, and training-based correctors~\citep{remedi2025,schiff2026proseco,kim2025prism,he2025mdpo,yao2026selfgen}. These methods act across steps by undoing or overwriting tokens whose semantics turn out to be wrong. In contrast, \ours{} acts within a step and reduces the factorization error before inconsistent tuples are written.

\paragraph{Joint prediction and iterated computation.}
MAR~\citep{li2024mar} models a block's joint distribution with a diffusion head on top of a transformer backbone. Stage~B is a lightweight instance of this idea: it uses one deterministic coordination pass with shared weights and a cross-entropy readout, applied online to the positions about to be committed rather than offline to an entire block. Iterating the coordination pass relates to weight-tied recurrent computation~\citep{dehghani2019universal,bai2019deq} and recurrent-depth test-time scaling~\citep{geiping2025recurrent}. The parameter $K$ controls coordination depth at commitment time and provides a test-time compute axis for parallel decoders. Classical iterated conditional modes~\citep{besag1986icm} provides a token-feedback analogue, operating on discrete tokens rather than hidden states.

%=============================================================================
\section{Preliminaries: LLaDA2.1 Training and Inference}
\label{sec:prelim}
%=============================================================================

We use LLaDA2.1~\citep{nie2026llada21} as the base system, a semi-autoregressive block-diffusion dLLM that generates text block by block, using iterative denoising within each block. This section reviews the base training and inference procedure used by LLaDA2.1.

\subsection{Base Training}
\label{sec:prelim-train}

\paragraph{Block-diffusion forward.}
The sequence is partitioned into blocks of $B$ tokens. Attention is block-causal. Within a block, every position, including \mask{} positions, attends to every other position \emph{bidirectionally}. Across blocks, attention is causal, so block $j$ sees only blocks $0, \dots, j$. Training and inference share this structure.

\paragraph{M2T/T2T supervision.}
Training corrupts a clean sequence $x = (x_1, \dots, x_T)$ into an input $z$ containing three kinds of positions: a masked set $\Mset$ (input \mask{}), a wrong-visible set $\Wset$ (visible tokens differing from the clean token), and a clean-visible set $\Gset$ (visible tokens equal to the clean token). The model reads the full corrupted input $z$ and outputs one token distribution $p_i$ per position; every supervised position is trained with cross-entropy toward its clean token:
\begin{equation}
    \mathcal{L}
    \;=\;
    \lambda_{\mathrm{m2t}} \!\sum_{i \in \Mset} \mathrm{CE}(p_i, x_i)
    \;+\;
    \lambda_{\mathrm{edit}} \!\sum_{i \in \Wset} \mathrm{CE}(p_i, x_i)
    \;+\;
    \lambda_{\mathrm{clean}} \!\sum_{i \in \Gset} \mathrm{CE}(p_i, x_i).
    \label{eq:threeterm}
\end{equation}
The three terms train mask filling (\MtoT{}), error editing (\TtoT{}-edit), and preservation of correct tokens (\TtoT{}-clean). They correspond to the two write operations used at inference.

\subsection{Base Inference}
\label{sec:prelim-infer}

\paragraph{Block-wise denoising.}
The response is generated block by block, left to right. Prompt tokens enter the sequence unmasked and are never edited. Once a response block is complete, its tokens are never edited again; only positions in the current block are initialized as \mask{} and updated by denoising.

\paragraph{Factorized M2T/T2T updates.}
The current block is decoded by iterative denoising. Let $\ctx$ denote the fixed context outside the current block, comprising the prompt and completed response blocks whose tokens are no longer edited. Let $z$ denote the current block state, including committed tokens and remaining \mask{} positions. Each step runs one forward pass through the $L$-layer backbone on $(\ctx, z)$, and the language-model head reads out one \emph{independent} distribution $p_\theta(\cdot \mid \ctx, z)$ per position. The step updates the block by two rules that share these logits. First, \emph{mask-to-token} (\MtoT{}) commitment selects a commit set $\cset$ from masked positions whose maximum probability exceeds $\tau_{\mathrm{m2t}}$. If none exceeds the threshold, it selects the single most confident position. Every $i \in \cset$ is then written to its argmax token independently:
\begin{equation}
    z_i \leftarrow \arg\max_{v \in \mathcal{V}} \, p_\theta(v \mid \ctx, z),
    \qquad i \in \cset .
    \label{eq:factorized}
\end{equation}
Second, \emph{token-to-token} (\TtoT{}) editing: every already-visible position of the current block is re-examined, and its current token $\tilde{z}_i$ is overwritten when a different candidate is sufficiently confident:
\begin{equation}
    z_i \leftarrow z^{*}_i
    \quad \text{if } p^{*}_i > \tau_{\mathrm{t2t}} \text{ and } z^{*}_i \neq \tilde{z}_i,
    \qquad
    z^{*}_i = \arg\max_{v \in \mathcal{V}} p_\theta(v \mid \ctx, z), \;\;
    p^{*}_i = \max_{v \in \mathcal{V}} p_\theta(v \mid \ctx, z).
    \label{eq:t2t}
\end{equation}
Steps repeat until the block contains no \mask{} and no edit fires (up to a maximum number of denoising steps). The completed block then joins the fixed context $\ctx$ and is never edited in later blocks. The base decoder only exposes per-position distributions $p_\theta(z_i \mid \ctx, z)$ from the language-model head. When several masked positions are selected in the same commit set $\cset$, the decoder does not choose their values jointly; it applies an independent argmax at each $i \in \cset$. The \TtoT{} edits in the same step are also read from the same pre-update state $z$: they can revise already-visible tokens, but they do not condition on the newly written \MtoT{} tokens from that step. Thus a single step has no explicit joint choice among simultaneous \MtoT{} writes, or between those writes and simultaneous \TtoT{} edits. We refer to this update rule as \emph{factorized commitment}.

%=============================================================================
\section{Method}
\label{sec:method}
%=============================================================================

\begin{figure}[t]
    \centering
    \includegraphics[width=\linewidth]{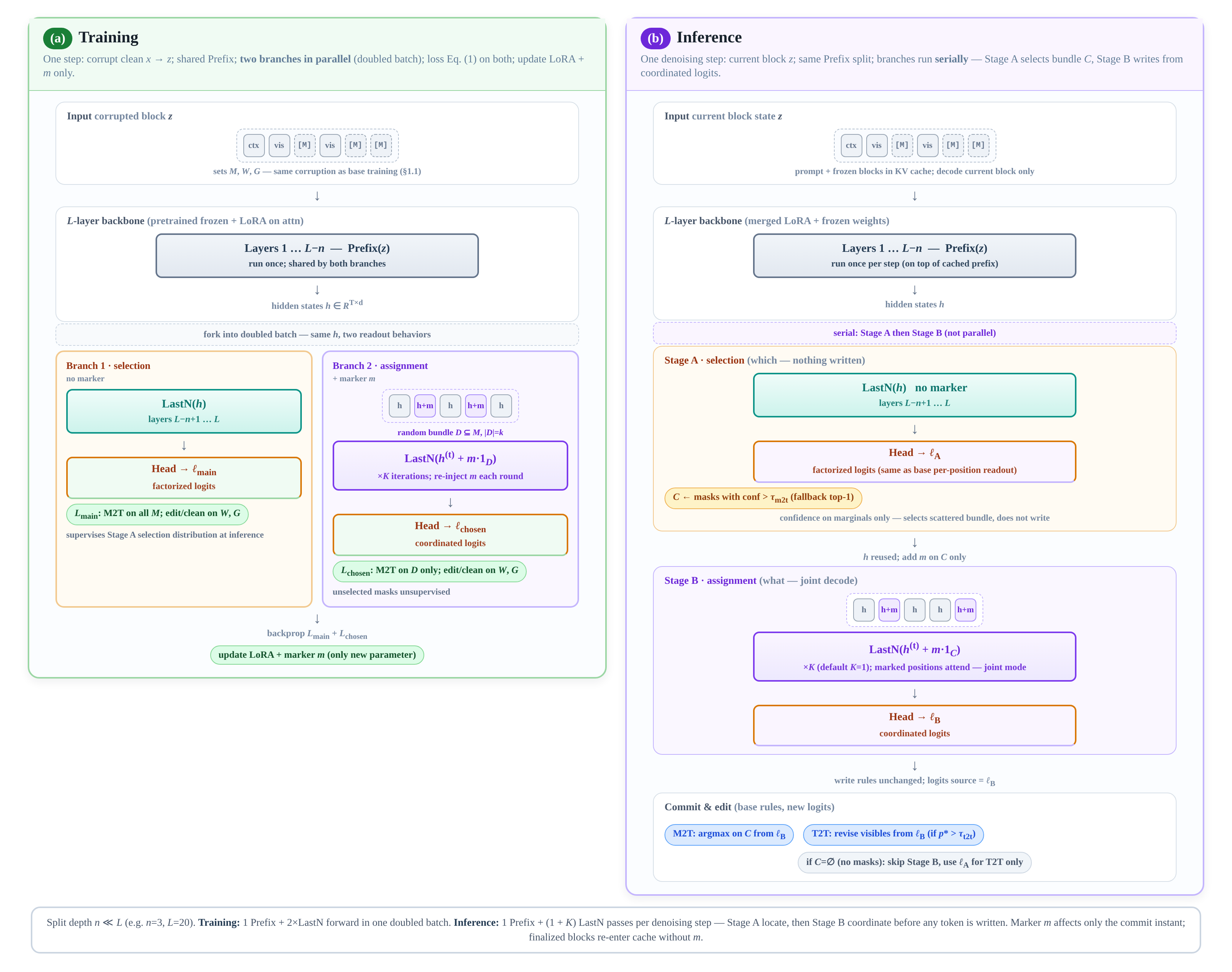}
    \caption{Split forward pass through the $L$-layer backbone.
    \textbf{(a)~Training:} a shared Prefix feeds two parallel branches: factorized selection without a marker and marked coordination on a random bundle.
    \textbf{(b)~Inference:} the same branches run serially. Stage~A selects the commit bundle, and Stage~B coordinates it before \MtoT{}/\TtoT{} writes.}
    \label{fig:architecture}
\end{figure}

LLaDA2.1's original decoding step is factorized. \ours{} replaces the original \MtoT{}/\TtoT{} update with a marker-gated coordination pass, turning the step's updates into a joint-mode commitment over the selected commit bundle. Training (Section~\ref{sec:training}) teaches the backbone this behavior through two branches; inference (Section~\ref{sec:inference}) runs the same two branches serially (Figure~\ref{fig:architecture}).

\paragraph{Compute-constrained validation.}
Ideally, marker-gated coordination would be integrated throughout pretraining, instruction tuning, and alignment so that the backbone learns to coordinate at every denoising stage. Full-scale training is beyond our available compute. We instead use a minimal implementation. Starting from the released LLaDA2.1-mini checkpoint, we freeze the backbone and train only low-rank LoRA adapters and marker vectors in a short continued-pretraining pass on generic English prose (Section~\ref{sec:setup}). This setting serves as a controlled test of the mechanism rather than a final training recipe. The results test whether a small adapter can learn split-forward coordination on a pretrained model using out-of-domain CPT text; they do not establish that this LoRA checkpoint is the best possible implementation.

\subsection{Training}
\label{sec:training}

Training applies the split forward of Figure~\ref{fig:architecture} to the corrupted sequence $z$ defined in Section~\ref{sec:prelim-train}. The two branches run in parallel on a doubled batch. We construct $z$ from clean $x$ using the self-generated $x \to z$ corruption of \citet{yao2026selfgen}. A no-gradient draft pass populates $\Wset$ with the model's own draft errors rather than the random vocabulary substitutions used in base LLaDA2.1 training.

Each example computes hidden states $h = \mathrm{Prefix}(z)$ once from the first $L - n$ layers; write $\mathrm{LastN}(\cdot)$ for the remaining $n$ layers.

\paragraph{Branch 1: selection branch (no marker).}
$h$ passes directly through the last $n$ layers, yielding $\ell_{\mathrm{main}} = \mathrm{Head}(\mathrm{LastN}(h))$. We apply the full three-term loss of Eq.~\eqref{eq:threeterm}, with \MtoT{} on all of $\Mset$ and edit and clean losses on $\Wset$ and $\Gset$. At inference, the selector uses these unmarked factorized logits to choose $\cset$. This branch therefore retains direct supervision for the distribution used by the selector.

\paragraph{Branch 2: assignment branch (marker).}
Branch~2 starts from $h = \mathrm{Prefix}(z)$.
(i)~Sample $|\dset| = s \sim \mathrm{Uniform}\{s_{\min}, \dots, s_{\max}\}$ and draw $\dset \subseteq \Mset$ uniformly, forming the position indicator $\mathbf{1}_\dset$ with $(\mathbf{1}_\dset)_j = 1$ iff $j \in \dset$.
(ii)~For $i = 0, \dots, K{-}1$,
\begin{equation}
    h^{(0)} = h, \qquad
    h^{(i+1)} = \mathrm{LastN}\!\left(h^{(i)} + \mathbf{1}_\dset \odot \marker_i\right), \qquad
    \ell_{\mathrm{chosen}} = \mathrm{Head}\!\left(h^{(K)}\right),
    \label{eq:iterate}
\end{equation}
where each $\marker_i$ is a learned marker vector for round~$i$.
(iii)~Apply Eq.~\eqref{eq:threeterm} to $\ell_{\mathrm{chosen}}$: edit and clean on all of $\Wset$ and $\Gset$, \MtoT{} on $\dset$ only.

Random bundles include the confidence-selected bundles encountered at inference as special cases, so training requires no confidence computation. Supervision uses ordinary cross-entropy against the ground-truth tokens. Coordination is induced by conditioning rather than by a joint loss: branch~2 predicts the bundle while observing which positions belong to it, and shared attention is the only path through which this information can affect the prediction. In implementation, the two branches are concatenated into one doubled batch and share one prefix forward and one backward pass.

\subsection{Inference}
\label{sec:inference}

Inference applies the same split forward to each denoising step, but runs the two branches serially. Branch~1 computes $h = \mathrm{Prefix}(\ctx, z)$ and factorized logits $\ell_A = \mathrm{Head}(\mathrm{LastN}(h))$, then selects $\cset$ using the rule in Section~\ref{sec:prelim-infer}. If $\cset \neq \varnothing$, branch~2 coordinates the selected bundle using Eq.~\eqref{eq:iterate} with $\mathbf{1}_\cset$ and sets $\ell = \mathrm{Head}(h^{(K)})$; otherwise, $\ell = \ell_A$. Both \MtoT{} and \TtoT{} write from $\ell$ according to Eqs.~\eqref{eq:factorized}--\eqref{eq:t2t}. Stage~B re-applies the last $n$ layers only to the current block and reuses the prefix key--value cache. It therefore adds approximately $n/L$ of one block forward per denoising step.

\paragraph{What Stage~B computes.}
Stage~A and Stage~B use the same per-position softmax but represent different conditionals. Stage~A estimates $p(x_i \mid \ctx, z)$, whose marginals average over the joint modes of $p(x_\cset \mid \ctx, z)$ in ambiguous contexts. Stage~B estimates $p(x_i \mid \ctx, z, \mathbf{1}_\cset)$, conditioning on the announcement that the positions in $\cset$ are being decided together. The announcement contains no token values because the marked positions remain \mask{}. Coordination occurs in hidden-state space: attention reveals to each marked position which other positions are being resolved in the same step, allowing their predictions to sharpen toward a compatible completion. Branch~2 makes this conditioning useful by supervising the entire bundle with cross-entropy toward one clean sequence. The coordinated predictions therefore co-adapt toward a single joint completion rather than a mixture of modes. Under greedy decoding, their per-position argmax tends to produce a tuple from one mode of $p(x_\cset \mid \ctx, z)$, yielding approximate joint-mode commitment even though each output remains a per-position conditional. This behavior is an inductive bias from the conditioning and training objective, not a structural guarantee; Section~\ref{sec:experiments} evaluates it empirically.

%=============================================================================
\section{Experiments}
\label{sec:experiments}
%=============================================================================

We compare the released LLaDA2.1-mini base model against our LoRA adapter trained with \ours{} on seven benchmarks under the shared protocol below (Table~\ref{tab:main}).

\subsection{Experimental Setup}
\label{sec:setup}

\paragraph{Model and inference.}
All runs use LLaDA2.1-mini (16B MoE, 20 transformer layers)~\citep{nie2026llada21}. Benchmark numbers use the official \texttt{generate()} path in Q~Mode with generation length 16384, 32 steps per block, block length $B{=}32$, $\tau_{\mathrm{m2t}}{=}0.7$, at most 16 post-steps, greedy decoding (temperature~0), and batch size~1. The base column uses the released checkpoint with factorized decoding at its released edit gate $\tau_{\mathrm{t2t}}{=}0.5$. \ours{} uses the LoRA and marker weights trained below and adds a marker-gated coordination pass over the last $n{=}3$ transformer layers with $K{=}1$ round (Section~\ref{sec:inference}); our default setting removes the confidence gate on \TtoT{} edits ($\tau_{\mathrm{t2t}}{=}0$), letting the coordinated Stage~B logits decide revisions directly. All other inference code and hyperparameters are shared between the two conditions.

\paragraph{Training.}
We start from the released LLaDA2.1-mini checkpoint, freeze the backbone, and train LoRA adapters ($r{=}16$, $\alpha{=}32$, Kaiming initialization on the attention query--key--value and output projections of all $L$ layers) together with $K$ zero-initialized marker vectors $\{\marker_0, \dots, \marker_{K-1}\}$. One marker is used per coordination round; these are the only parameters added beyond LoRA. The training set contains 50k FineWeb-Edu documents with quality score $\ge 3$ and 256--1800 tokens per document. This generic English prose is unrelated to the evaluation benchmarks. We construct corrupted inputs $z$ from clean sequences $x$ following \citet{yao2026selfgen} and add the marker branch from Section~\ref{sec:training}. Branch~2 marks a random bundle of $s$ positions, where $s \sim \mathrm{Uniform}\{1, \dots, 16\}$. We optimize with AdamW at learning rate $1{\times}10^{-5}$ for two epochs (global batch size~8, sequence length~2048, cosine schedule with $1\%$ warmup) under FSDP2 with bf16 and gradient checkpointing. The loss weights are $(\lambda_{\mathrm{m2t}}, \lambda_{\mathrm{edit}}, \lambda_{\mathrm{clean}}) = (1.0, 0.3, 0.2)$, and we evaluate the checkpoint at training step~12500.

\paragraph{Benchmarks.}
The reported set covers seven benchmarks grouped into four categories.
GSM8K is a grade-school math word-problem benchmark with numeric final answers, scored by exact match~\citep{cobbe2021gsm8k}.
CMATH is a Chinese elementary-school math benchmark with short numeric final answers, scored by numeric exact match~\citep{wei2023cmath}.
GPQA-Diamond is a graduate-level science multiple-choice benchmark, scored by accuracy~\citep{rein2024gpqa}.
DROP is a reading-comprehension benchmark requiring discrete reasoning over paragraphs, scored by exact match on the validation split~\citep{dua2019drop}.
PIQA is a physical-commonsense two-choice benchmark, scored by accuracy~\citep{bisk2020piqa}.
HumanEval+ and MBPP+ are the EvalPlus-hardened versions of the HumanEval and MBPP code-generation benchmarks, scored by pass@1 with the official EvalPlus harness~\citep{chen2021humaneval,austin2021mbpp,liu2023evalplus}.

\subsection{Main Results}
\label{sec:exp-main}

\begin{table}
\centering
\caption{Main results on 7 benchmarks. Both columns use LLaDA2.1-mini under the shared protocol of Section~\ref{sec:setup} (base: released factorized decoding with $\tau_{\mathrm{t2t}}{=}0.5$; \ours{}: marker-gated coordination with ungated edits, $\tau_{\mathrm{t2t}}{=}0$). Metrics are accuracy, exact match, or pass@1 (\%); best results are in \textbf{bold}. NFE/tok.\ is base\,/\,\ours{} full-network forwards per generated token; for \ours{} the Stage-B partial pass over the last $n{=}3$ layers is not counted (it adds ${\approx}0.15\times$ per coordinated step).}
\label{tab:main}
\vspace{4pt}
\small
\renewcommand{\arraystretch}{1.0}
\begin{tabular}{llcccc}
\toprule
\textbf{Category} & \textbf{Benchmark} & \textbf{Base} & \textbf{\ours{}} & \textbf{$\Delta$} & \textbf{NFE/tok.} \\
\midrule
\multirow{2}{*}{Math}
  & GSM8K          & 89.61          & \textbf{91.13} & $+1.52$ & 0.26 / 0.34 \\
  & CMATH          & \textbf{92.26} & 91.99          & $-0.27$ & 0.32 / 0.45 \\
\midrule
Science
  & GPQA-Diamond   & 43.94          & \textbf{46.97} & $+3.03$ & 0.41 / 0.35 \\
\midrule
\multirow{2}{*}{Reasoning}
  & DROP (EM)      & \textbf{84.14} & 83.93          & $-0.21$ & 0.33 / 0.29 \\
  & PIQA           & 80.90          & \textbf{84.60} & $+3.70$ & 0.76 / 0.44 \\
\midrule
\multirow{2}{*}{Code}
  & HumanEval+     & 76.83          & \textbf{82.32} & $+5.49$ & 0.14 / 0.16 \\
  & MBPP+          & 68.78          & \textbf{70.11} & $+1.33$ & 0.18 / 0.18 \\
\bottomrule
\end{tabular}
\end{table}

Table~\ref{tab:main} shows that \ours{} improves code generation (HumanEval+ $+5.49$, MBPP+ $+1.33$), physical commonsense (PIQA $+3.70$), graduate-level science QA (GPQA $+3.03$), and grade-school math (GSM8K $+1.52$). CMATH ($-0.27$) and DROP ($-0.21$) remain nearly unchanged. The largest gains occur on tasks where several simultaneously committed positions must agree on a coherent completion, including code and multi-token exact answers. This pattern is consistent with the within-step coordination error that \ours{} targets. Marker-gated coordination reduces factorization error at commitment time instead of relying on downstream repair. The forward counts place both decoders in a similar budget range. \ours{} uses fewer full-network forwards on PIQA ($0.44$ vs.\ $0.76$), GPQA ($0.35$ vs.\ $0.41$), and DROP ($0.29$ vs.\ $0.33$), matches MBPP+, and uses moderately more on GSM8K, CMATH, and HumanEval+.

\subsection{Ablations}
\label{sec:exp-ablation}

The ablations address two questions. First, which component accounts for the gain: the adapters, the coordination pass, the logits used for writing, or the thresholds? We study this question on all $1319$ GSM8K problems (Table~\ref{tab:abl-mech}). Second, does coordination benefit from re-applying more layers? We vary last $n$ on all seven benchmarks (Table~\ref{tab:abl-lastn}).

\begin{table}
\centering
\caption{Mechanism ablation on GSM8K ($n{=}1319$ per cell). $\Delta$ is relative to the \ours{} default ($91.13$). NFE/tok counts full-network forwards per generated token; the Stage-B partial pass over the last $n{=}3$ layers is not included (it adds ${\approx}0.15\times$ the counted cost per coordinated step). ``Fill'' is which logits commit masks (\MtoT{}), ``edit'' is which logits revise visible tokens (\TtoT{}).}
\label{tab:abl-mech}
\vspace{4pt}
\small
\renewcommand{\arraystretch}{1.05}
\begin{tabular}{lcccrr}
\toprule
\textbf{Variant} & $\tau_{\mathrm{m2t}}$ & $\tau_{\mathrm{t2t}}$ & \textbf{Acc} & \textbf{$\Delta$} & \textbf{NFE/tok} \\
\midrule
\multicolumn{6}{l}{\emph{Full systems}} \\
Released, factorized decoding & 0.7 & 0.5 & 89.61 & $-1.52$ & 0.26 \\
Our weights, factorized decoding (Stage~B off) & 0.7 & 0.5 & 90.75 & $-0.38$ & 0.28 \\
\ours{}, gated edits & 0.7 & 0.5 & \textbf{91.13} & $0.00$ & 0.28 \\
\ours{} (default) & 0.7 & 0 & \textbf{91.13} & --- & 0.34 \\
\midrule
\multicolumn{6}{l}{\emph{Which logits write (our weights, Stage~B on)}} \\
Stage~B fills, Stage~A edits & 0.7 & 0.5 & 90.90 & $-0.23$ & 0.27 \\
Stage~A fills, Stage~B edits & 0.7 & 0.5 & 90.90 & $-0.23$ & 0.30 \\
Stage~B fills, editing disabled & 0.7 & --- & 88.40 & $-2.73$ & 0.30 \\
Last-$n$ re-applied \emph{without} marker & 0.7 & 0.5 & 90.37 & $-0.76$ & 0.28 \\
\midrule
\multicolumn{6}{l}{\emph{Commit selection}} \\
Entropy-based selection (matched $|\cset|$) & --- & 0.5 & 90.30 & $-0.83$ & 0.27 \\
\ours{}, lower commit threshold & 0.5 & 0.5 & 88.40 & $-2.73$ & 0.24 \\
\ours{}, lower commit threshold & 0.3 & 0.5 & 84.00 & $-7.13$ & 0.22 \\
Our weights, Stage~B off & 0.5 & 0.5 & 88.48 & $-2.65$ & 0.23 \\
Our weights, Stage~B off & 0.3 & 0.5 & 83.70 & $-7.43$ & 0.22 \\
Released & 0.5 & 0.5 & 86.73 & $-4.40$ & 0.26 \\
Released & 0.3 & 0.5 & 79.15 & $-11.98$ & 0.30 \\
\midrule
\multicolumn{6}{l}{\emph{Maximum-parallelism references}} \\
Released, both gates removed & 0 & 0 & 83.40 & $-7.73$ & 0.15 \\
All-marker (mark every mask, commit all) & --- & 0 & 79.91 & $-11.22$ & 0.10 \\
\bottomrule
\end{tabular}
\end{table}

\paragraph{Decomposing the gain.}
The \emph{full systems} block separates the effect of the trained weights from the coordination mechanism. With Stage~B disabled, our weights improve by $1.14$ points over the released model, capturing the effect of the adapters and CPT data. Enabling Stage~B at the same weights adds another $0.38$ points. The \emph{write-source} block further shows that coordinated logits should control both operations. Routing either fills or edits to the factorized Stage~A logits reduces accuracy by $0.23$ points, while disabling editing reduces it by $2.73$ points. Re-applying the last $n$ layers without a marker reduces accuracy by $0.76$ points, indicating that the marker announcement contributes beyond additional depth. Across the seven benchmarks, removing the edit gate under coordination ($\tau_{\mathrm{t2t}}\!: 0.5 \to 0$) improves GPQA ($+1.52$), MBPP+ ($+1.06$), HumanEval+ ($+0.61$), DROP ($+0.44$), and PIQA ($+0.11$), leaves GSM8K unchanged, and reduces CMATH by $0.45$ points. These results support using the coordinated logits directly for revision.

\paragraph{Commit selection remains important.}
At matched bundle sizes, entropy-based selection reduces accuracy by $0.83$ points. Confidence therefore remains the stronger commit criterion, consistent with our claim that its limitation is missing dependency information rather than poor calibration. Lowering the commit threshold degrades every system. The trained weights, rather than Stage~B, reduce the sensitivity: at $\tau_{\mathrm{m2t}}{=}0.3$, the released model loses $10.46$ points from its default, whereas both of our variants lose approximately $7.1$ points. Coordination therefore does not compensate for indiscriminate commitment. Removing both gates from the released decoder ($83.40$) and marking every mask ($79.91$) also perform poorly. However, their NFE values ($0.15$ and $0.10$) indicate that these settings commit $2$--$3\times$ more tokens per forward. Their degradation thus conflates reduced selectivity with reduced coordination, so we report them as reference settings rather than controlled ablations.

\begin{table}
\centering
\caption{Coordination-depth ablation: last-$n$ layers re-applied in Stage~B, trained and evaluated at matching $n$ (all with $\tau_{\mathrm{t2t}}{=}0$). Best accuracy per row in \textbf{bold}. NFE/tok.\ is given as $n{=}3\,/\,6\,/\,9$ and counts full-network forwards only; the uncounted Stage-B partial pass grows with $n$ (${\approx}n/L$ per coordinated step), so equal counted NFE means larger $n$ is strictly more expensive.}
\label{tab:abl-lastn}
\vspace{4pt}
\small
\renewcommand{\arraystretch}{1.0}
\begin{tabular}{lcccc}
\toprule
\textbf{Benchmark} & $n{=}3$ (default) & $n{=}6$ & $n{=}9$ & \textbf{NFE/tok.} \\
\midrule
GSM8K        & \textbf{91.13} & 90.75 & 90.83 & 0.34 / 0.30 / 0.30 \\
CMATH        & 91.99          & \textbf{93.08} & 92.17 & 0.45 / 0.44 / 0.40 \\
GPQA-Diamond & \textbf{46.97} & 42.93 & 39.39 & 0.35 / 0.36 / 0.35 \\
DROP         & 83.93          & 82.80 & \textbf{84.32} & 0.29 / 0.32 / 0.35 \\
PIQA         & 84.60          & 83.68 & \textbf{84.77} & 0.44 / 0.47 / 0.38 \\
HumanEval+   & \textbf{82.32} & 80.49 & 79.88 & 0.16 / 0.14 / 0.14 \\
MBPP+        & 70.11          & 68.78 & \textbf{72.22} & 0.18 / 0.18 / 0.18 \\
\bottomrule
\end{tabular}
\end{table}

\paragraph{Three layers suffice.}
Table~\ref{tab:abl-lastn} varies the depth of the coordination pass. Averaged over the seven benchmarks, $n{=}3$ scores $78.72$, compared with $77.50$ for $n{=}6$ and $77.65$ for $n{=}9$. No benchmark improves monotonically with depth, and GPQA decreases from $46.97$ to $42.93$ and $39.39$. Although $n{=}9$ obtains the best result on DROP, PIQA, and MBPP+, tripling coordination depth provides no consistent benefit. This result suggests that a short communication path is sufficient to resolve the bundle mode, consistent with joint structure already being present in the prefix hidden states. We adopt $n{=}3$ as the default because it has the highest average score and the lowest cost.

%=============================================================================
\section{Conclusion}
\label{sec:conclusion}
%=============================================================================

Parallel commitment in diffusion language models incurs a per-step distributional error given by the conditional total correlation of the commit bundle. Confidence-based selection cannot observe this error because per-position marginals average over joint modes and contain no dependency signal. \ours{} addresses the error at commitment time. A learned marker identifies the bundle, and the backbone's last $n$ layers are re-applied so the selected positions coordinate before token writes. Under greedy commitment, one deterministic coordination round approximates joint-mode argmax in place of independent marginal argmax. The added components are lightweight adapters, one marker vector, and a partial forward pass. Two extensions remain open: marking visible tokens could unify commitment and revision, while iterating the coordination pass could trade test-time compute for stronger within-bundle dependence. Within-step coordination does not eliminate the need for across-step repair because a coordinated bundle can still be incorrect. It instead reduces the factorization error that repair methods otherwise address only after commitment.

\paragraph{Limitations.}
\ours{} requires fine-tuning (adapters and the marker); it is not a training-free rule. The coordination claim is scoped to greedy commitment: under temperature sampling, per-position noise can still split a bundle across modes unless the coordinated marginals are sharp. The hidden-state iteration for $K > 1$ carries no theoretical monotonicity guarantee, and our checkpoints are trained at $K = 1$, so deeper coordination requires training with sampled $K$. Marking visible tokens for joint revision is discussed but not evaluated here. Finally, all experiments use one model family (LLaDA2.1-mini); the generality of the mode-averaging diagnosis across dLLM families remains to be established.

\bibliographystyle{plainnat}
\bibliography{refs}

\end{document}